\documentclass[letterpaper]{article}
\usepackage[submission]{aaai2027}
\usepackage[hyphens]{url}
\usepackage{graphicx}
\usepackage{natbib}
\usepackage{caption}
\usepackage{algorithm}
\usepackage{algorithmic}
\usepackage{multirow}
\usepackage{newfloat}
\usepackage{listings}
\DeclareCaptionStyle{ruled}{labelfont=normalfont,labelsep=colon,strut=off}
\floatstyle{ruled}
\newfloat{listing}{tb}{lst}{}
\floatname{listing}{Listing}
\usepackage{booktabs}
\usepackage{amsmath}
\usepackage{amssymb}
\usepackage{mathtools}
\makeatletter
\def\showauthors@on{T}
\makeatother
\title{PhiFold: Towards Dynamic Protein Design with Physics-Structured Covariance Modeling}
\author{
    Yutian Liu\textsuperscript{\rm 1}\equalcontrib,
    Mujie Lin\textsuperscript{\rm 2}\equalcontrib,
    Lanqian Zhang\textsuperscript{\rm 3}\equalcontrib,
    Meng Fan\textsuperscript{\rm 2},
    Chang Liu\textsuperscript{\rm 4}\corresponding,
    Zhiwei Nie\textsuperscript{\rm 5}\corresponding,
    Siwei Ma\textsuperscript{\rm 1}\corresponding
}
\affiliations{
    \textsuperscript{\rm 1}School of Computer Science, Peking University, Beijing, China\\
    \textsuperscript{\rm 2}School of Electronic and Computer Engineering,
    Peking University, Shenzhen, China\\
    \textsuperscript{\rm 3}School of Life Sciences, Tsinghua University, Beijing 100084, China\\
    \textsuperscript{\rm 4}Department of Automation and BNRist,
    Tsinghua University, Beijing, China\\
    \textsuperscript{\rm 5}Institute of Science and Technology for Brain-Inspired Intelligence, Fudan University, Shanghai, China\\
}
\begin{document}
\maketitle
\begin{abstract}
Protein design is moving beyond structural correctness toward function-aware design, yet existing generative models typically treat dynamics as a downstream property estimated through simulation or prediction after structure generation. Using MD trajectories as a generative target is also undesirable because stochastic, path-dependent trajectories over-specify the underlying equilibrium ensemble. We introduce PhiFold, a framework for jointly generating protein backbones and their second-order dynamics, represented by residue-displacement covariance. Rather than predicting the quadratically sized full covariance, PhiFold decomposes dynamics into three interpretable components: local flexibility, a low-rank collective-motion representation, and residue-wise collective participation. These components are assembled into a positive-definite covariance matrix with exact marginal consistency, yielding a compact and physically constrained representation of equilibrium dynamics. Across generated proteins, PhiFold improves recovery of local fluctuations and long-range residue coupling while remaining competitive on dominant collective-motion subspaces. It further enables bidirectional control of residue flexibility
while preserving backbone designability. By unifying structure generation with an explicit representation of equilibrium dynamics, PhiFold lays a foundation for designing proteins not only by how they look, but also by how they move.
\end{abstract}
\section{Introduction}
\begin{figure*}[t]
\centering
\includegraphics[width=2.0\columnwidth]{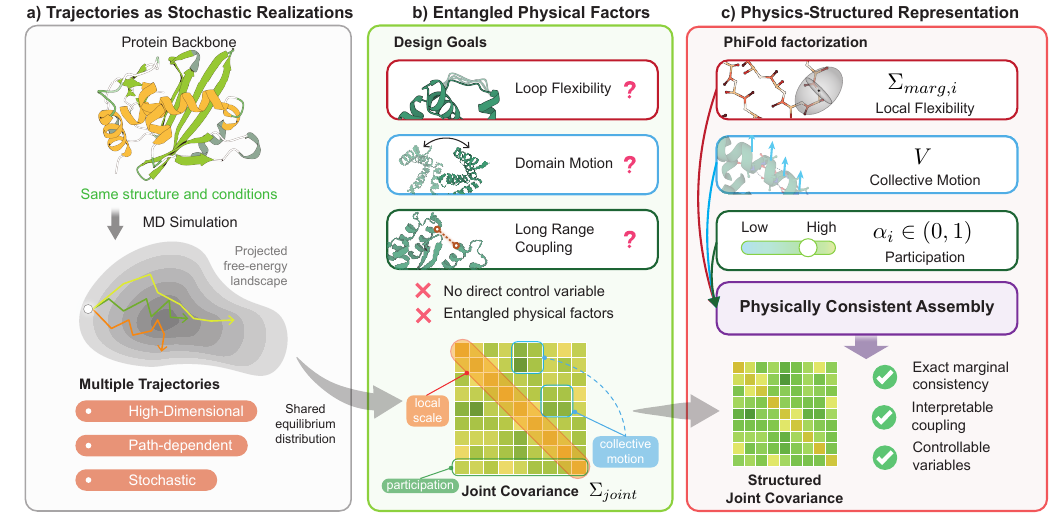}
\caption{
\textbf{Motivation and overview of PhiFold.}
MD trajectories are high-dimensional, path-dependent, and stochastic. PhiFold instead models equilibrium covariance through physically interpretable factors for local flexibility, collective motion, and residue-wise participation.
}
\label{fig1}
\end{figure*}
Recent advances in generative modeling have enabled high-quality
\textit{de novo} protein design~\cite{watson2023denovo,yim2023se,bose2024se3}.
However, existing methods largely generate static structures, while
conformational dynamics are addressed only through post hoc prediction or
molecular simulation. Since dynamics are essential for protein
function~\cite{karplus2005molecular,henzlerwildman2007dynamic}, we aim to
develop a generative framework that jointly models protein structures and
intrinsic dynamics.
A natural approach to incorporating dynamics into generative protein design is
to model molecular dynamics (MD) trajectories directly. However, an MD
trajectory is a stochastic and path-dependent realization sampled from an
underlying equilibrium ensemble. Under identical thermodynamic
conditions, different initial velocities, thermostats, or random seeds
can produce distinct trajectories that represent the same equilibrium
distribution. Trajectory-level supervision therefore over-specifies the
learning target by requiring models to capture trajectory-specific
variations rather than shared equilibrium properties. A more appropriate
representation is the second-order statistics of the equilibrium ensemble.
After removing rigid-body motions, residue-displacement covariance
captures equilibrium fluctuations and inter-residue correlations
~\cite{ichiye1991collective}, while its leading eigenspace describes
dominant collective motions~\cite{amadei1993essential,hayward1995collective,
david2014pca}. We therefore formulate protein dynamics generation as the
joint generation of protein backbones and their equilibrium covariance.
However, directly predicting the full covariance remains challenging:
its dimensionality grows quadratically with protein length, and the
matrix entangles dynamical effects across different physical scales,
from local residue fluctuations to collective motions. Such a
heterogeneous representation creates a difficult-to-interpret learning
objective and limits explicit control over individual dynamical
properties.
To address these limitations, we introduce a physics-structured
representation of equilibrium covariance. Rather than treating covariance
as a monolithic matrix, we decompose it into three physically
interpretable components: local flexibility for residue-wise fluctuations,
collective geometry for coherent motions, and collective participation for
residue contributions to collective modes. This decomposition separates
distinct dynamical factors and enables their controlled assembly into a
valid covariance representation. Together, these components capture the
dominant local and collective information in equilibrium covariance with
low empirical reconstruction error (cf. Fig.~\ref{fig1}).
Building on this representation, we propose PhiFold, a unified
framework that couples protein backbone generation with dynamics modeling.
PhiFold predicts the three physical components and generates a valid
covariance representation, enabling interpretable and controllable
dynamics generation. Coupled with a flow-based backbone generator,
PhiFold supports joint structure--dynamics design and explicit control
over regional flexibility.
We evaluate PhiFold by comparing its predicted covariance against all-atom GROMACS simulations and existing
dynamics models, including post hoc approaches MDGen~\cite{jing2024generative}
and ANM~\cite{atilgan2001anisotropy}, as well as conformational sampling
models BioEmu~\cite{lewis2025bioemu} and Str2Str~\cite{lu2024str2str}.
PhiFold achieves the best recovery of local fluctuation statistics and
residue-level dynamical coupling while remaining competitive on collective
motion metrics. Ablation studies demonstrate that the physics-structured
decomposition is essential for covariance validity and predictive
accuracy, and controllable generation experiments demonstrate targeted
regional flexibility modulation without compromising backbone
designability.
Our contributions are:
(i)~a formulation of backbone-conditioned, near-equilibrium dynamics
generation through joint backbone--covariance modeling, using residue-
displacement covariance as a time-order-independent second-order summary
of equilibrium fluctuations;
(ii)~a physics-structured covariance parameterization that separates
local flexibility, collective geometry, and residue-wise participation,
while guaranteeing positive definiteness and exact marginal consistency;
and
(iii)~PhiFold, which predicts these quantities from shared backbone-flow
representations, enables gradient-based regional flexibility control
during backbone sampling, and is validated on generated proteins against
independent all-atom MD simulations and diverse post hoc baselines.
\section{Related Work}
\subsection{Generative Protein Design}
Inverse-folding models, such as ProteinMPNN~\cite{dauparas2022robust} and
ESM-IF~\cite{hsu2022learning}, predict sequences conditioned on fixed
backbones, while generative backbone models including RFdiffusion
~\cite{watson2023denovo}, FrameDiff~\cite{yim2023se}, Chroma
~\cite{ingraham2023chroma}, FoldFlow~\cite{bose2024se3}, and SCUBA-D
~\cite{liu2024scubad} generate novel structures or sequence--structure
distributions. Despite their success in static protein design, these
methods generally do not explicitly model equilibrium dynamics.
Recent efforts incorporate dynamics through multi-state design,
programmable conformational constraints, or predefined motion descriptors.
DynamicMPNN~\cite{abrudan2026dynamicmpnn} and SwitchCraft
~\cite{jing2026switchcraft} design sequences under multiple or
state-dependent conformations, while dynamics-informed conditioning,
NMA-tune~\cite{komorowska2025nmatune}, and VibeGen~\cite{ni2025vibegen}
leverage prescribed motion patterns. However, these approaches rely on
selected conformations or predefined modes rather than generating the
equilibrium dynamics associated with each designed backbone.
PhiFold differs by generating protein backbones with their
equilibrium dynamics. Instead of representing motion through discrete
states or selected modes, PhiFold models residue-displacement covariance
as a compact representation of anisotropic fluctuations, residue
couplings, and collective motions, enabling unified and controllable
structure--dynamics design.
\subsection{Protein Dynamics Modeling}
MD simulations provide atomistic trajectories but are computationally expensive and sampling-dependent. Harmonic models provide efficient approximations of near-native motions, with Gaussian network models capturing isotropic fluctuations and residue correlations, and anisotropic network models modeling directional collective motions~\cite{bahar1997gnm,atilgan2001anisotropy}.
Recent models generate conformational ensembles conditioned on existing
sequences or structures~\cite{jing2024alphaflow,zheng2024distributional,
wang2024confdiff,lewis2025bioemu}. Temporal approaches instead learn
accelerated transition proposals~\cite{klein2023timewarp} or
structure-conditioned trajectories~\cite{jing2024generative}.
Concurrent DynaProt~\cite{bafna2026learningresiduelevelprotein} predicts
multiscale Gaussian dynamics for a fixed structure.
PhiFold addresses a different task by jointly generating backbones and equilibrium dynamics. Rather than deriving dynamics from sampled conformations or trajectories, PhiFold directly predicts ensemble covariance as a compact representation of equilibrium motion.
\section{Method}
\begin{figure*}[t]
\centering
\includegraphics[width=2.0\columnwidth]{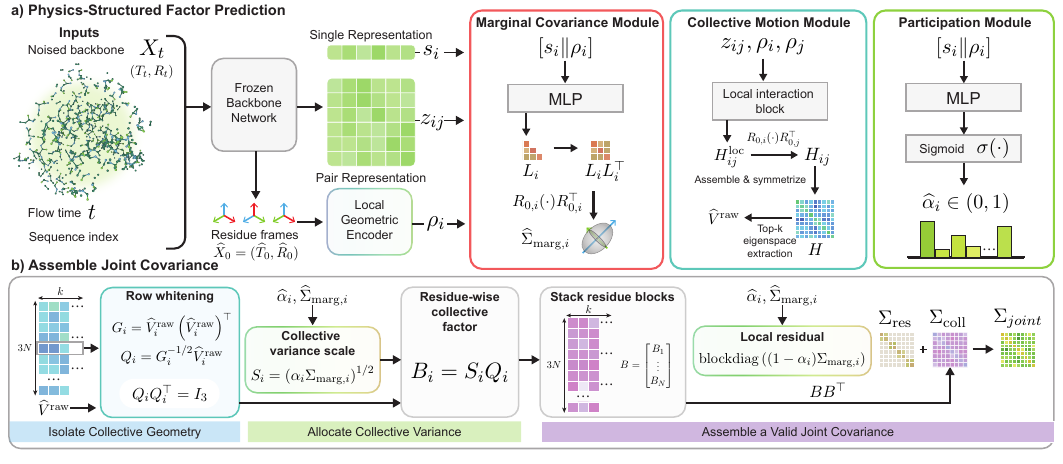}
\caption{
\textbf{Physics-structured covariance modeling in PhiFold.}
Three prediction modules estimate marginal covariance, collective geometry, and participation. These factors are normalized, variance-calibrated, and assembled into a valid joint covariance with exact marginal consistency.
}
\label{fig2}
\end{figure*}
\subsection{Problem Setup}
\label{sec:problem_setup}
For a protein of $N$ residues, we represent the backbone atoms
$(\mathrm{N},\mathrm{C}_{\alpha},\mathrm{C})$ of each residue as a
local rigid frame following AlphaFold2~\cite{jumper2021highly}. The
backbone is parameterized as
\begin{equation}
X=(T,R)\in SE(3)^N,
\label{eq:backbone_representation}
\end{equation}
where $T\in\mathbb{R}^{N\times3}$ and $R\in SO(3)^N$ denote the
residue-wise translations and rotations, respectively.
We seek to model the joint distribution
$
p(X,\mathcal{E}),
$
where $\mathcal{E}$ denotes the equilibrium ensemble associated with
backbone $X$ under fixed thermodynamic conditions. Rather than
representing $\mathcal{E}$ by a MD trajectory, we
characterize its second-order equilibrium statistics through residue
displacements. After rigid-body alignment, let
$\delta x_i\in\mathbb{R}^3$ denote the displacement of residue $i$
from its equilibrium mean, and define
\begin{equation}
\delta X
=
\left[
\delta x_1^\top,\ldots,\delta x_N^\top
\right]^\top
\in\mathbb{R}^{3N}.
\label{eq:stacked_displacement}
\end{equation}
The equilibrium displacement covariance, estimated empirically from
MD trajectories, is
\begin{equation}
\Sigma_{\mathrm{emp}}
=
\mathbb{E}_{\mathcal{E}}
\left[
\delta X\delta X^\top
\right]
\in\mathbb{S}_{+}^{3N}.
\label{eq:empirical_covariance}
\end{equation}
Its $(i,j)$-th $3\times3$ block describes the correlated Cartesian
fluctuations between residues $i$ and $j$. Because rigid-body
alignment removes global translational and rotational degrees of
freedom, $\Sigma_{\mathrm{emp}}$ is generally positive semidefinite
and may contain null modes.
Under a local Gaussian approximation, we treat $X$ as the equilibrium
mean structure and represent $\mathcal{E}$ by its displacement
covariance. We define the generation target as
\begin{equation}
\left(
X,\widehat{\Sigma}_{\mathrm{joint}}
\right),
\qquad
\widehat{\Sigma}_{\mathrm{joint}}
\in
\mathbb{S}_{++}^{3N},
\label{eq:generation_target}
\end{equation}
where $\widehat{\Sigma}_{\mathrm{joint}}$ is a structured
positive-definite approximation to the aligned empirical covariance
$\Sigma_{\mathrm{emp}}$. This representation captures the internal
second-order fluctuation statistics while regularizing
alignment-induced null directions.
\subsection{Equilibrium Covariance Modeling}
\label{sec:covariance_modeling}
We express the equilibrium covariance
$\Sigma_{\mathrm{joint}}$ in residue-wise block form:
\begin{equation}
\Sigma_{\mathrm{joint}}
=
\begin{bmatrix}
\Sigma_{11} & \cdots & \Sigma_{1N}\\
\vdots & \ddots & \vdots\\
\Sigma_{N1} & \cdots & \Sigma_{NN}
\end{bmatrix},
\qquad
\Sigma_{ij}\in\mathbb{R}^{3\times3}.
\label{eq:covariance_blocks}
\end{equation}
The diagonal block
$
\Sigma_{\mathrm{marg},i}
=
\Sigma_{ii}
\in\mathbb{S}_{++}^{3}
$
captures the magnitude and anisotropy of residue $i$'s fluctuations,
whereas each off-diagonal block $\Sigma_{ij}$ characterizes correlated
motion between residues $i$ and $j$. We denote the collection of
residue-wise marginal covariances by
$
\Sigma_{\mathrm{marg}}
=
\{\Sigma_{\mathrm{marg},i}\}_{i=1}^{N}.
$
Although $\Sigma_{\mathrm{joint}}$ provides a compact second-order
description of equilibrium dynamics, directly parameterizing the full
matrix remains challenging: the number of entries grows quadratically
with protein length, and distinct dynamical factors are entangled
within a monolithic representation. This limits interpretability and
provides no explicit interface for controlling individual dynamical
properties. We therefore introduce a physically structured
parameterization that separates local flexibility, a collective-motion
representation, and residue-wise collective participation. These
components are subsequently assembled into a positive-definite
covariance with exact residue-wise marginal consistency.
\subsubsection{Dynamics Decomposition}
\label{sec:dynamics_decomposition}
We parameterize the equilibrium covariance using three complementary
components:
\begin{equation}
\left\{
\Sigma_{\mathrm{marg}},
V^{\mathrm{raw}},
\alpha
\right\},
\label{eq:dynamics_variables}
\end{equation}
where $\Sigma_{\mathrm{marg}}$ describes residue-wise marginal
fluctuations, $V^{\mathrm{raw}}\in\mathbb{R}^{3N\times K}$
is a raw low-dimensional representation of collective motion, and
$\alpha\in(0,1)^N$ specifies residue-wise collective participation
fractions. Here, $V^{\mathrm{raw}}$ is an intermediate representation
that encodes collective-motion geometry and does not necessarily
coincide with the leading eigenspace of the final assembled covariance.
Together, these components define a variance-calibrated collective
factor
\begin{equation}
B
=
\mathcal{B}
\left(
V^{\mathrm{raw}},
\Sigma_{\mathrm{marg}},
\alpha
\right)
\in\mathbb{R}^{3N\times K},
\label{eq:collective_factor_mapping}
\end{equation}
which induces a collective covariance $BB^\top$ of rank at most $K$.
This representation separates the geometry of collective motion,
encoded by $V^{\mathrm{raw}}$, from the residue-wise allocation of
fluctuation variance, controlled by $\alpha$. We next describe the
variance calibration and covariance assembly.
\subsubsection{Covariance Assembly}
\label{sec:covariance_assembly}
As shown in Fig.~\ref{fig2}b, we construct the collective factor $B$ such that the $i$-th diagonal
block of the induced collective covariance satisfies
$[BB^\top]_{ii}=\alpha_i\Sigma_{\mathrm{marg},i}$.
Let $V_i^{\mathrm{raw}}\in\mathbb{R}^{3\times K}$ denote the row
block of $V^{\mathrm{raw}}$ associated with residue $i$. All matrix
square roots and inverse square roots below denote principal symmetric
roots. We first normalize each residue-wise block:
\begin{equation}
G_i
=
V_i^{\mathrm{raw}}
\left(V_i^{\mathrm{raw}}\right)^\top,
\qquad
Q_i
=
G_i^{-1/2}V_i^{\mathrm{raw}}.
\label{eq:block_whitening}
\end{equation}
Assuming $K\geq3$ and $V_i^{\mathrm{raw}}$ has full row rank,
$G_i\succ0$ and $Q_iQ_i^\top=I_3$. Thus, $Q_i$ preserves the row
space of $V_i^{\mathrm{raw}}$ while removing its residue-wise scale.
We then compute the symmetric square root of the marginal covariance
allocated to collective motion:
\begin{equation}
S_i
=
\left(
\alpha_i\Sigma_{\mathrm{marg},i}
\right)^{1/2},
\label{eq:symmetric_sqrt}
\end{equation}
and define the residue-wise collective factor as
\begin{equation}
B_i
=
S_iQ_i
\in\mathbb{R}^{3\times K}.
\label{eq:collective_factor}
\end{equation}
After stacking the residue-wise blocks into
$B\in\mathbb{R}^{3N\times K}$, we assemble the joint covariance as
\begin{equation}
\begin{aligned}
\Sigma_{\mathrm{joint}}
&=
BB^\top+\Sigma_{\mathrm{res}},\\
\Sigma_{\mathrm{res}}
&=
\operatorname{blockdiag}
\left(
\Sigma_{\mathrm{res},1},
\ldots,
\Sigma_{\mathrm{res},N}
\right),\\
\Sigma_{\mathrm{res},i}
&=
(1-\alpha_i)\Sigma_{\mathrm{marg},i}.
\end{aligned}
\label{eq:joint_assembly}
\end{equation}
The low-rank term $BB^\top$ captures coordinated cross-residue
fluctuations, while the block-diagonal residual models residue-local
motion. Thus, $\alpha_i$ controls the allocation of residue $i$'s marginal
covariance between collective and local components.
The following properties hold under the full-row-rank assumption
($\mathrm{rank}(V_i^{\mathrm{raw}})=3$). Eigenvalue flooring is used for
numerical stability in ill-conditioned cases; however, it was not required across all evaluated proteins, where all
residue-wise blocks remained full rank (see Supplementary).
\subsubsection{Theoretical Properties}
\label{sec:ensemble_analysis}
\paragraph{Exact marginal consistency.}
Under the full-row-rank assumption in
Eq.~\eqref{eq:block_whitening}, the $i$-th diagonal block of the
collective covariance satisfies
\begin{equation}
\left[BB^\top\right]_{ii}
=
B_iB_i^\top
=
S_iQ_iQ_i^\top S_i^\top
=
\alpha_i\Sigma_{\mathrm{marg},i}.
\label{eq:collective_diagonal}
\end{equation}
Combining the collective and residue-local components gives
\begin{equation}
\begin{aligned}
[\Sigma_{\mathrm{joint}}]_{ii}
&=
\alpha_i\Sigma_{\mathrm{marg},i}
+
(1-\alpha_i)\Sigma_{\mathrm{marg},i}
=
\Sigma_{\mathrm{marg},i}.
\end{aligned}
\label{eq:exact_diagonal}
\end{equation}
Thus, the assembled covariance exactly preserves the prescribed
residue-wise marginal covariances. This consistency is guaranteed by
construction rather than enforced through an additional objective.
\paragraph{Positive definiteness.}
For $\Sigma_{\mathrm{marg},i}\succ0$ and $\alpha_i\in(0,1)$, each
residual block satisfies
$
\Sigma_{\mathrm{res},i}
=
(1-\alpha_i)\Sigma_{\mathrm{marg},i}
\succ0,
$
and therefore $\Sigma_{\mathrm{res}}\succ0$. For any nonzero
$z\in\mathbb{R}^{3N}$,
\begin{equation}
\begin{aligned}
z^\top\Sigma_{\mathrm{joint}}z
&=
z^\top BB^\top z
+
z^\top\Sigma_{\mathrm{res}}z\\
&=
\lVert B^\top z\rVert_2^2
+
z^\top\Sigma_{\mathrm{res}}z
>
0.
\end{aligned}
\label{eq:positive_definiteness}
\end{equation}
Hence, $\Sigma_{\mathrm{joint}}\succ0$.
\paragraph{Factorized residue coupling.}
For two distinct residues $i\neq j$, the off-diagonal covariance
block is
\begin{align}
\Sigma_{ij}
&=
B_iB_j^\top
=
S_iQ_iQ_j^\top S_j^\top
\nonumber\\
&=
\sqrt{\alpha_i\alpha_j}\,
\Sigma_{\mathrm{marg},i}^{1/2}
Q_iQ_j^\top
\Sigma_{\mathrm{marg},j}^{1/2}.
\label{eq:factorized_coupling}
\end{align}
This factorization separates cross-residue coupling into the local
marginal covariances, the collective alignment $Q_iQ_j^\top$, and the
participation strength $\sqrt{\alpha_i\alpha_j}$.
To remove the local fluctuation scales, we define the normalized
coupling block
\begin{equation}
C_{ij}
=
\Sigma_{\mathrm{marg},i}^{-1/2}
\Sigma_{ij}
\Sigma_{\mathrm{marg},j}^{-1/2}
=
\sqrt{\alpha_i\alpha_j}\,
Q_iQ_j^\top.
\label{eq:normalized_coupling}
\end{equation}
Since $Q_iQ_i^\top=Q_jQ_j^\top=I_3$, we have
$\lVert Q_i\rVert_2=\lVert Q_j\rVert_2=1$, and therefore
\begin{equation}
\lVert C_{ij}\rVert_2
\leq
\sqrt{\alpha_i\alpha_j}.
\label{eq:coupling_ceiling}
\end{equation}
Thus, the collective participation fractions provide an explicit
upper bound on normalized cross-residue coupling.
\paragraph{Collective-basis invariance.}
For any orthogonal matrix $O\in O(K)$, replacing
$V_i^{\mathrm{raw}}$ by $V_i^{\mathrm{raw}}O$ leaves $G_i$ unchanged
and transforms $Q_i$ and $B_i$ as
$
Q_i\mapsto Q_iO
$
and
$
B_i\mapsto B_iO.
$
Consequently,
\begin{equation}
(BO)(BO)^\top
=
BB^\top,
\end{equation}
so the assembled covariance is independent of the particular
orthonormal basis used to represent the collective subspace.
\paragraph{Rotation equivariance.}
Under a global rotation $U\in SO(3)$, the inputs transform as
$
V_i^{\mathrm{raw}}\mapsto UV_i^{\mathrm{raw}}
$
and
$
\Sigma_{\mathrm{marg},i}
\mapsto
U\Sigma_{\mathrm{marg},i}U^\top.
$
The assembly consequently gives $B_i\mapsto UB_i$, and hence
\begin{equation}
\Sigma_{\mathrm{joint}}
\mapsto
(I_N\otimes U)
\Sigma_{\mathrm{joint}}
(I_N\otimes U)^\top.
\label{eq:covariance_equivariance}
\end{equation}
The resulting covariance is therefore $SO(3)$-equivariant and
consistent across Cartesian coordinate frames.
The proposed parameterization represents covariance within a structured
low-rank-plus-block-diagonal family. On ATLAS, the leading $K=20$
collective modes capture $95.8\%$ of the total variance, while oracle
reconstruction within this family achieves a mean relative Frobenius
error of $0.037$. Additional analyses are provided in the Supplement.
\subsection{Coupled Backbone--Dynamics Generation}
\label{sec:coupled_generation}
Unlike sequential pipelines that generate a static backbone before
estimating its dynamics through MD or a separate predictor, our
framework generates backbone geometry and equilibrium dynamics from
shared flow representations(Fig.~\ref{fig2}a).  Given a noised backbone
$X_t=(T_t,R_t)\in SE(3)^N$, a pretrained $SE(3)$-equivariant flow
network predicts the denoised endpoint
$\widehat X_0=(\widehat T_0,\widehat R_0)$ and exposes residue-wise and
pairwise intermediate features
\begin{equation}
s_i\in\mathbb{R}^{d_s},
\qquad
z_{ij}\in\mathbb{R}^{d_p}.
\label{eq:shared_flow_features}
\end{equation}
Throughout the dynamics branch, the predicted endpoint rotations
$\widehat R_{0,i}$ serve as local reference frames for mapping
dynamical quantities to the global coordinate system of the generated
backbone. The dynamics branch directly uses these intermediate features rather
than operating only on the final generated structure.
We further encode the local geometric environment of residue $i$ as
\begin{equation}
\rho_i
=
f_{\mathrm{ctx}}
\left(
s_i,\{\phi_{ij}\}_{j\in\mathcal N_i}
\right),
\label{eq:rigidity_context}
\end{equation}
where $\phi_{ij}$ contains distance, relative position, relative
orientation, and sequence-separation features expressed in
residue-local frames. The resulting context $\rho_i$ is used by the
dynamics prediction modules.
The marginal covariance is predicted through a positive-diagonal
triangular factor:
\begin{equation}
\begin{aligned}
L_i
&=
\operatorname{Tri}_{+}
\left(
f_{\mathrm{marg}}(s_i,\rho_i)
\right),\\
\widehat\Sigma_{\mathrm{marg},i}
&=
R_{0,i}L_iL_i^\top R_{0,i}^\top
\end{aligned}
\label{eq:marginal_prediction}
\end{equation}
where $\operatorname{Tri}_{+}$ maps the diagonal entries through a
strictly positive transformation.
This construction guarantees
$\widehat\Sigma_{\mathrm{marg},i}\in\mathbb S_{++}^{3}$ and preserves
rotational equivariance.
To model collective motion, pairwise features are mapped to
residue-local interaction blocks:
\begin{equation}
H_{ij}^{\mathrm{loc}}
=
f_{\mathrm{pair}}
\left(
z_{ij},\rho_i,\rho_j
\right),
\qquad
H_{ij}
=
 R_{0,i}
H_{ij}^{\mathrm{loc}}
R_{0,j}^{\top}.
\label{eq:interaction_blocks}
\end{equation}
We assemble these blocks into
$H_{\mathrm{raw}}\in\mathbb{R}^{3N\times3N}$ and construct the
symmetric global interaction operator
\begin{equation}
H
=
\frac{1}{2}
\left(
H_{\mathrm{raw}}+H_{\mathrm{raw}}^\top
\right).
\label{eq:interaction_symmetrization}
\end{equation}
The eigenvectors corresponding to the $K$ largest algebraic
eigenvalues of $H$ define the raw collective-motion representation
\begin{equation}
\widehat V^{\mathrm{raw}}
=
\operatorname{EigVec}_K(H).
\label{eq:collective_basis_prediction}
\end{equation}
An independent branch predicts the collective participation fraction
\begin{equation}
\widehat\alpha_i
=
\sigma
\left(
f_\alpha(s_i,\rho_i)
\right).
\label{eq:alpha_prediction}
\end{equation}
The resulting
$\widehat\Sigma_{\mathrm{marg}}$,
$\widehat V^{\mathrm{raw}}$, and
$\widehat\alpha$ are combined by the constrained assembly in
Sec.~\ref{sec:covariance_assembly}.
The pretrained backbone-flow parameters remain fixed, while the dynamics module retains differentiable access to the noised backbone state, enabling dynamics-guided backbone generation.
The backbone generator follows the original FoldFlow objective. The dynamics branch is trained to recover MD-derived equilibrium statistics, including marginal fluctuations, collective motion subspaces, and residue coupling. Detailed objectives are described in the Supplement.
\section{Experiments}
\subsection{Experiments Setup}
\begin{table*}[t]
\centering
\setlength{\tabcolsep}{4.5pt}
\begin{tabular*}{\textwidth}{
ll
ccccc
|c
@{}
}
\toprule
Aspect
& Metric
& PhiFold
& MDGen
& ANM
& BioEmu
& Str2Str
& GROMACS \\
&
& \textit{Joint}
& \textit{Post hoc}
& \textit{Post hoc}
& \textit{Post hoc}
& \textit{Post hoc}
& \textit{R1--R2} \\
\midrule
\multicolumn{8}{l}{
\textbf{A. Fine-grained covariance fidelity}
} \\
\addlinespace[1pt]
\multirow{3}{*}{Local fluctuation}
& RMS displacement $r$ $\uparrow$
& $\mathbf{0.739}$
& $0.516$
& $\underline{0.663}$
& $0.492$
& $0.480$
& $0.795$ \\
& RMS displacement NMAE $\downarrow$
& $\mathbf{0.303}$
& $\underline{0.563}$
& $0.838$
& $3.932$
& $4.210$
& $0.241$ \\
& Marginal eigenvalue $r$ $\uparrow$
& $\mathbf{0.682}$
& $0.577$
& $\underline{0.608}$
& $0.536$
& $0.566$
& $0.769$ \\
\addlinespace[1pt]
Collective motion
& Top-20 subspace overlap $\uparrow$
& $\underline{0.428}$
& $\mathbf{0.520}$
& $0.427$
& $0.356$
& $0.328$
& $0.695$ \\
\addlinespace[1pt]
\multirow{2}{*}{Residue coupling}
& DCCM $r$ $\uparrow$
& $\underline{0.358}$
& $0.173$
& $\mathbf{0.476}$
& $0.351$
& $0.344$
& $0.591$ \\
& Long-range DCCM $r$ $\uparrow$
& $0.204$
& $0.066$
& $\mathbf{0.338}$
& $\underline{0.209}$
& $0.179$
& $0.549$ \\
\midrule
\multicolumn{8}{l}{
\textbf{B. Ensemble-level fidelity}
} \\
\addlinespace[1pt]
Local fluctuation
& Per-target RMSF $r$ $\uparrow$
& $\mathbf{0.811}$
& $\underline{0.758}$
& $0.724$
& $0.651$
& $0.553$
& $0.836$ \\
\addlinespace[1pt]
\multirow{2}{*}{Distributional accuracy}
& Root-mean $W_2$ $\downarrow$
& $\mathbf{1.386}$
& $\underline{1.484}$
& $1.505$
& $6.905$
& $7.047$
& $1.025$ \\
& Variance $W_2$ $\downarrow$
& $\mathbf{0.518}$
& $0.806$
& $\underline{0.795}$
& $5.158$
& $5.961$
& $0.380$ \\
\addlinespace[1pt]
Collective motion
& MD-PCA $W_2$ $\downarrow$
& $\underline{10.599}$
& $12.473$
& $\mathbf{9.863}$
& $27.995$
& $17.625$
& $8.654$ \\
\addlinespace[1pt]
\multirow{2}{*}{Ensemble observables}
& Weak contacts $J$ $\uparrow$
& $\mathbf{0.521}$
& $\underline{0.431}$
& $0.225$
& $0.425$
& $0.390$
& $0.692$ \\
& Transient contacts $J$ $\uparrow$
& $\mathbf{0.326}$
& $\underline{0.250}$
& $0.122$
& $0.193$
& $0.161$
& $0.586$ \\
\bottomrule
\end{tabular*}
\caption{
Agreement with independent two-replica all-atom GROMACS simulations on
the same set of generated proteins.
Fine-grained covariance diagnostics compare PhiFold with the
representative post hoc estimators MDGen and ANM, while unified
ensemble-level metrics compare all methods.
All entries report mean performance, and Wasserstein distances are in \AA.
GROMACS R1--R2 provides an empirical reference for finite-sampling
variability and is excluded from method ranking.
Best and second-best method results within each evaluation block are shown
in \textbf{bold} and \underline{underlined}, respectively.
Additional metrics are reported in the Supplement.
}
\label{tab:gromacs_comparison}
\end{table*}
\paragraph{Dataset.}
We train PhiFold on ATLAS~\cite{vandermeersche2024atlas}, comprising 1,390 protein chains with three independent 100-ns all-atom MD trajectories each. ATLAS is split at the protein-chain level into train/validation/test sets (1111/139/140 proteins), with all replicas of each protein kept in the same split to prevent leakage.
\paragraph{Model and training.}
PhiFold is implemented in PyTorch based on the FoldFlow architecture~\cite{bose2024se3}.
The backbone generation module is initialized from a FoldFlow checkpoint pretrained on 22,248 monomeric PDB structures, while the covariance prediction modules are trained from scratch.
During training, the backbone trunk is frozen and only the dynamics heads are optimized, preventing disruption of the pretrained structure generation capability.
The model is trained with Adam using a constant learning rate of $2\times10^{-4}$ on a NVIDIA H100 GPU.
Small eigenvalue floors are applied during dynamics decomposition for numerical stability.
\paragraph{Evaluation setup.}
Following the FoldFlow evaluation protocol~\cite{bose2024se3}, we generate
protein backbones across lengths
$\{100,150,200,250,300\}$. For each backbone, eight sequences are designed
with ProteinMPNN~\cite{dauparas2022robust} and refolded with
ESMFold~\cite{lin2023evolutionary}. Designable proteins are selected using
structure recovery metrics
($\mathrm{scRMSD}<2\,\text{\AA}$ and $\mathrm{TMscore}>0.5$); no
dynamics-related metric is used for selection. PhiFold generates covariance
together with the backbone and is not rerun after sequence design and
refolding.
We compare PhiFold with post hoc dynamics models, including
MDGen~\cite{jing2024generative}, ANM~\cite{atilgan2001anisotropy},
BioEmu~\cite{lewis2025bioemu}, and Str2Str~\cite{lu2024str2str}, using the same generated backbones and designed
sequences. MDGen generates structure-conditioned trajectories, while ANM
models collective motions through an elastic network. All methods are
evaluated against independent all-atom GROMACS simulations following the
ATLAS protocol~\cite{vandermeersche2024atlas} with the CHARMM36m force
field. For atomistic validation, 250 generated candidates passing the
designability criteria are randomly subsampled to 50 proteins with
uniform length coverage for independent GROMACS simulations.
\subsection{Main Results}
\label{sec:dynamics_fidelity}
We evaluate 50 randomly selected designable proteins, uniformly covering
lengths $\{100,150,200,250,300\}$, against two-replica all-atom GROMACS
simulations. All methods are evaluated on the same proteins using
fine-grained covariance diagnostics and unified ensemble-level metrics.
\paragraph{Fine-grained covariance fidelity.}
As shown in Table~\ref{tab:gromacs_comparison}A, PhiFold leads all three
local-fluctuation metrics, demonstrating accurate recovery of both
fluctuation magnitude and directional variance.
MDGen achieves the highest top-20 subspace overlap, while ANM performs best
on DCCM-based coupling, consistent with their respective strengths in
trajectory-derived collective motion and elastic-network correlations.
PhiFold nevertheless remains competitive on both aspects, ranking second
in overall DCCM and closely matching BioEmu on long-range coupling.
\paragraph{Ensemble-level fidelity.}
More importantly, Table~\ref{tab:gromacs_comparison}B shows that PhiFold
ranks first on five of six unified metrics: per-target RMSF, overall and
variance Wasserstein errors, and weak- and transient-contact agreement.
It also ranks second to ANM on MD-PCA distance.
These results show that PhiFold's covariance accuracy translates into
well-calibrated ensemble spread and residue-contact statistics rather than
improvements confined to individual covariance diagnostics.
Overall, PhiFold leads eight of the twelve reported metrics and provides
the strongest cross-scale fidelity, while specialized post hoc methods
retain advantages on particular collective-motion measures.
Notably, PhiFold achieves this performance by jointly predicting dynamics
with the backbone, before the final sequence-realized structure available
to post hoc methods.
The supplementary analysis further shows that its ensembles can be
generated efficiently through direct covariance sampling.
\subsection{Ablation Study}
\label{sec:ablation}
\begin{table*}[t]
\centering
\small
\setlength{\tabcolsep}{5.0pt}
\begin{tabular}{l|cc|ccc|c}
\toprule
Method
& \shortstack{Marginal eig.\\$r$\ $\uparrow$}
& \shortstack{Top-$20$ subspace\\overlap $\uparrow$}
& \shortstack{DCCM\\$r$\ $\uparrow$}
& \shortstack{Nonlocal DCCM\\$r$\ $\uparrow$}
& \shortstack{Long-range DCCM\\$r$\ $\uparrow$}
& \shortstack{Valid covariance\\(\%) $\uparrow$} \\
& \multicolumn{2}{c}{(MDGen proxy)}
& \multicolumn{3}{c}{(ANM proxy)}
& \\
\midrule
Direct joint
& $0.316 \pm 0.133$
& $0.311 \pm 0.073$
& $0.404 \pm 0.048$
& $0.163 \pm 0.048$
& $0.062 \pm 0.065$
& $\mathit{0.0}$ \\
Fixed $\alpha_i=0.5$
& $0.499 \pm 0.122$
& $0.414 \pm 0.059$
& $0.457 \pm 0.116$
& $0.379 \pm 0.090$
& $0.345 \pm 0.082$
& $\mathbf{100.0}$ \\
Cholesky
& $0.484 \pm 0.105$
& $0.383 \pm 0.051$
& $0.448 \pm 0.056$
& $0.273 \pm 0.046$
& $0.191 \pm 0.060$
& $\mathbf{100.0}$ \\
\textbf{Full}
& $\mathbf{0.531 \pm 0.131}$
& $\mathbf{0.432 \pm 0.052}$
& $\mathbf{0.513 \pm 0.100}$
& $\mathbf{0.423 \pm 0.086}$
& $\mathbf{0.386 \pm 0.086}$
& $\mathbf{100.0}$ \\
\bottomrule
\end{tabular}
\caption{
Ablation of the physics-structured covariance parameterization.
Values are mean $\pm$ standard deviation over 50 generated proteins.
Best proxy-recovery results are shown in \textbf{bold}.
}
\label{tab:ablation}
\end{table*}
We compare PhiFold with three variants under the same training and sampling
protocol.
\textit{Direct joint} unconstrainedly regresses the full covariance;
\textit{Cholesky} guarantees positive definiteness without the proposed
physical decomposition; and \textit{Fixed participation} sets
$\alpha_i=0.5$ for all residues.
Each variant is evaluated on 50 generated proteins evenly distributed
across lengths $\{100,150,200,250,300\}$.
Because repeating multi-replica GROMACS simulations for every variant is
prohibitively expensive, we use MDGen and ANM as proxies for collective
subspaces and residue coupling, respectively, following their strengths
established in Section~\ref{sec:dynamics_fidelity}.
These proxies are used only for architectural comparison; absolute
atomistic fidelity is evaluated in the main GROMACS experiment.
As shown in Table~\ref{tab:ablation}, direct joint regression produces no
positive-definite covariances and substantially degrades all recovery
metrics.
Cholesky restores covariance validity but remains consistently inferior
to the full model, showing that positive definiteness alone is
insufficient.
Fixing $\alpha_i=0.5$ causes smaller but systematic reductions,
particularly in residue-coupling recovery.
Together, these results support both the physics-structured decomposition
and residue-adaptive collective participation.
\subsection{Controllable Generation of Regional Flexibility}
\label{sec:regional_guidance}
\begin{figure}[t]
\centering
\includegraphics[width=0.99\columnwidth]{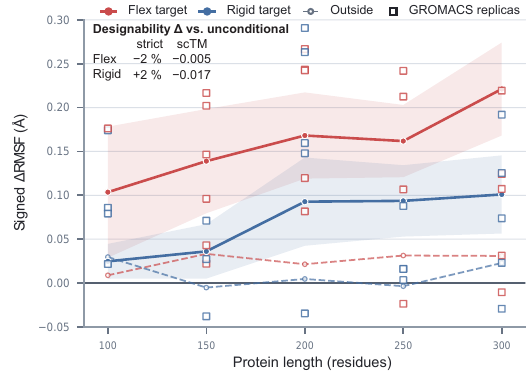}
\caption{
Controllable generation of regional flexibility.
Solid and dashed curves show signed target- and outside-region $\Delta$RMSF, with positive values indicating the intended direction; shading denotes 95\% confidence intervals over 20 matched samples.
Open squares show individual GROMACS replicas, and designability changes relative to unconditional generation are reported.
}
\label{fig:regional_guidance}
\end{figure}
Because the dynamics branch remains differentiably coupled to the backbone generator, its predicted regional fluctuations can guide backbone formation rather than modify the covariance post hoc. At each sampling step, we backpropagate the target-region fluctuation to the noisy backbone coordinates and use the gradient to modify the flow velocity. We increase fluctuations in helix-anchored regions and suppress them in coil-anchored regions. Guidance is applied over $t\in[0.5,0.9]$ with a scale of $0.5$. At each length, we generate 20 matched flexible, rigid, and unconditional triplets from shared initial noise, using an increased sampling noise scale of $0.5$ to evaluate control under greater structural diversity.
As shown in Figure~\ref{fig:regional_guidance}, both conditions consistently shift the generated target-region fluctuations in the intended direction. Flexibility guidance increases the target-region RMSF by $0.104$--$0.221$~\AA, whereas rigidity guidance decreases it by $0.025$--$0.101$~\AA. Outside-region changes generally remain within $0.033$~\AA, demonstrating spatially localized control. Independent GROMACS replicas exhibit greater variability, but a majority retain the intended direction under both conditions, with clearer agreement for flexibility guidance and greater length-dependent heterogeneity for rigidity guidance.
Backbone-level analyses further show that the control signal is encoded in the generated geometry: regional displacement, secondary-structure, and ANM measurements change preferentially within the target region while largely preserving global collective-motion organization. Guidance also preserves backbone quality, with aggregated designability changes within five percentage points and mean scTM changes within $0.017$. Together, these results demonstrate localized control of backbone-encoded dynamics without substantial degradation of designability (see Supplementary Information for details).
\section{Conclusion}
We introduced PhiFold, a framework for jointly generating de novo
protein backbones and backbone-conditioned residue-displacement
covariances.
By making equilibrium dynamics an explicit design variable, PhiFold moves
beyond post hoc dynamics estimation and integrates backbone generation,
ensemble modeling, and regional flexibility control within a unified
framework.
Its physics-structured representation separates local flexibility,
collective motion, and residue-wise participation while ensuring valid
covariances with exact marginal consistency.
PhiFold is particularly suited to early-stage design involving
near-equilibrium fluctuations, domain-scale collective motion, and
long-range dynamical coupling.
This scope is supported by biophysical evidence that protein contact
topology encodes a substantial fraction of slow collective dynamics and
often predisposes proteins toward functional motions
~\cite{ming2002describe,bahar2010global}.
However, topology does not fully determine the conformational energy
landscape: amino-acid identity, side-chain interactions, energetic
frustration, and environmental conditions can reshape state populations
and large-amplitude transitions~\cite{li2011frustration}.
PhiFold therefore provides a backbone-level foundation for dynamics-aware
design rather than a replacement for sequence-resolved atomistic modeling.
Extending it toward joint sequence--backbone--dynamics generation and
multimodal, temporally resolved ensembles is a natural direction for
future work.
\bibliography{aaai2027}
\clearpage
\section*{Supplementary Information}
\appendix
\section{Training Objectives}
\label{sec:training_objectives}
Let $\Sigma_{\mathrm{marg}}^{*}$ and $U_K^{*}$ denote the marginal
covariances and leading $K$ eigenvectors extracted from an MD
ensemble. The dynamics loss is
\begin{align}
\mathcal L_{\mathrm{dyn}}
={}&
\lambda_{\mathrm{marg}}\mathcal L_{\mathrm{marg}}
+\lambda_{\mathrm{sub}}\mathcal L_{\mathrm{sub}}
+\lambda_{\alpha}\mathcal L_{\alpha}
\nonumber\\
&+
\lambda_{\mathrm{frob}}\mathcal L_{\mathrm{frob}}
+\lambda_{\mathrm{pair}}\mathcal L_{\mathrm{pair}}.
\label{eq:total_dynamics_loss}
\end{align}
We supervise the marginal covariance using
\begin{equation}
\mathcal L_{\mathrm{marg}}
=
\frac{1}{N_{\mathrm{act}}^{\mathrm{eff}}}
\sum_i
m_i^{\mathrm{eff}}\,
d_{\mathrm{LC}}^2
\left(
\widehat\Sigma_{\mathrm{marg},i},
\Sigma_{\mathrm{marg},i}^{*}
\right).
\label{eq:marginal_loss}
\end{equation}
Here,
\begin{equation}
m_i
=
\mathrm{res\_mask}_i
\left(1-\mathrm{fixed\_mask}_i\right),
\qquad
N_{\mathrm{act}}=\sum_i m_i ,
\label{eq:active_mask}
\end{equation}
where $\mathrm{res\_mask}_i$ excludes padding residues and
$\mathrm{fixed\_mask}_i$ identifies residues excluded from the flow.
For the marginal loss, residues for which Cholesky decomposition
fails are additionally excluded:
\begin{equation}
m_i^{\mathrm{eff}}
=
m_i\,\mathbf{1}\!\left[\mathrm{chol\_ok}_i\right].
\qquad
N_{\mathrm{act}}^{\mathrm{eff}}
=
\sum_i m_i^{\mathrm{eff}} .
\end{equation}
For a symmetric positive-definite matrix $\Sigma$, let
$L=\operatorname{chol}(\Sigma)$ and let
$\lfloor L\rfloor$ denote its strictly lower-triangular part.
We define
\begin{align}
d_{\mathrm{LC}}^2(\Sigma_p,\Sigma_g)
={}&
\left\|
\lfloor L_p\rfloor-\lfloor L_g\rfloor
\right\|_F^2
\nonumber\\
&+
\left\|
\log\operatorname{diag}(L_p)
-
\log\operatorname{diag}(L_g)
\right\|_2^2 .
\label{eq:log_cholesky_distance}
\end{align}
In implementation, Cholesky diagonal entries are clamped below at
$10^{-8}$, and each per-residue loss is clamped above at $25$.
An auxiliary basis-invariant loss supervises the raw collective
representation:
\begin{equation}
\mathcal L_{\mathrm{sub}}
=
1-\frac{1}{K}
\left\|
(\widehat V^{\mathrm{raw}})^\top U_K^{*}
\right\|_F^2.
\label{eq:subspace_loss}
\end{equation}
For residue $i$, the PCA participation target is
\begin{equation}
\alpha_i^{*}
=
\frac{
\sum_{\ell=1}^{K}
\lambda_\ell^{*}
\|u_{i,\ell}^{*}\|_2^2
}{
\operatorname{tr}
(\Sigma_{\mathrm{marg},i}^{*})
},
\label{eq:alpha_target}
\end{equation}
where $u_{i,\ell}^{*}\in\mathbb R^3$ is the corresponding residue
block. We optimize
\begin{equation}
\mathcal L_{\alpha}
=
\frac{1}{N_{\mathrm{act}}}
\sum_i m_i\,
\operatorname{SmoothL1}
\left(
\widehat\alpha_i,
\bar\alpha_i^{*}
\right),
\label{eq:alpha_loss}
\end{equation}
where
\begin{equation}
\bar\alpha_i^{*}
=
\operatorname{clip}(\alpha_i^{*},0.20,0.99).
\end{equation}
For covariance-level supervision, we use the same reference
marginals on both sides to isolate collective supervision. The
masked Frobenius loss is
\begin{equation}
\mathcal L_{\mathrm{frob}}
=
\frac{
\left\|
M\odot
\left(
\widetilde\Sigma_{\mathrm{joint}}
-
\Sigma_{\mathrm{joint}}^{\mathrm{ref}}
\right)
\right\|_F^2
}{
9N_{\mathrm{act}}^2
}.
\label{eq:joint_frobenius_loss}
\end{equation}
To define the mask, we repeat each residue mask three times:
\begin{equation}
\widetilde m=m\otimes\mathbf 1_3,
\qquad
M_{ab}=\widetilde m_a\widetilde m_b .
\label{eq:joint_covariance_mask}
\end{equation}
Thus, a $3\times3$ block $(i,j)$ is valid if and only if both
residues are active. The mask covers the full $3N\times3N$
covariance matrix, consistent with the $N_{\mathrm{act}}^2$
normalization. The reference covariance
$\Sigma_{\mathrm{joint}}^{\mathrm{ref}}$ is reconstructed from the
empirical collective modes and reference marginal covariances.
Finally, $\mathcal L_{\mathrm{pair}}$ supervises local off-diagonal
covariance blocks. For each residue $i$, let
$\mathcal N_i^{(K_{\mathrm{nn}})}$ contain its
$K_{\mathrm{nn}}=8$ nearest spatial neighbors according to
ground-truth $C_\alpha$ distances. Define
\begin{equation}
B_{ij}
=
[\widetilde\Sigma_{\mathrm{joint}}]_{ij}
\in\mathbb R^{3\times3},
\end{equation}
and let $B_{ij}^{\mathrm{ref}}$ be assembled from the empirical
collective modes and PCA-derived participation targets
$\bar\alpha^{*}$. The pairwise loss is
\begin{align}
\mathcal L_{\mathrm{pair}}
={}&
\frac{1}{9N_{\mathrm{pair}}}
\sum_i
\sum_{j\in\mathcal N_i^{(K_{\mathrm{nn}})}}
m_i m_j
\left\|
B_{ij}-B_{ij}^{\mathrm{ref}}
\right\|_F^2,
\label{eq:pair_loss}\\
N_{\mathrm{pair}}
={}&
\sum_i
\sum_{j\in\mathcal N_i^{(K_{\mathrm{nn}})}}
m_i m_j .
\end{align}
Because the covariance assembly preserves the marginal blocks by
construction, no additional marginal-consistency loss is required.
\begin{table}[htbp]
\centering
\small
\begin{tabular}{ccc}
\hline
Symbol & Description & Value \\
\hline
$\lambda_{\mathrm{marg}}$
& Marginal covariance loss weight
& $1.0$ \\
$\lambda_{\mathrm{sub}}$
& Subspace alignment loss weight
& $1.0$ \\
$\lambda_{\alpha}$
& PCA participation regularization weight
& $0.3$ \\
$\lambda_{\mathrm{frob}}$
& Joint covariance Frobenius loss weight
& $0.05$ \\
$\lambda_{\mathrm{pair}}$
& Local pairwise covariance loss weight
& $1.0$ \\
$K_{\mathrm{nn}}$
& Number of local spatial neighbors
& $8$ \\
\hline
\end{tabular}
\caption{Hyperparameter settings.}
\label{tab:loss_weights_n15b_sfm_fix}
\end{table}
\section{Evaluation Metrics}
\label{sec:eval_metrics}
The computational definitions of the metrics reported in the paper are as follows.
\paragraph{Local flexibility.}
Let $\Sigma_{ii} \in \mathbb{R}^{3 \times 3}$ denote the marginal displacement covariance of residue $i$.
We define its displacement magnitude as
\begin{equation}
    f_i = \sqrt{\operatorname{tr}(\Sigma_{ii})}.
\end{equation}
Local agreement is evaluated using the Pearson correlation between the predicted and simulated $\{f_i\}_{i=1}^{N}$, together with the normalized mean absolute error
\begin{equation}
    E_{\mathrm{flex}}
    =
    \frac{
        \frac{1}{N}\sum_{i=1}^{N}
        \left|\hat{f}_i-f_i^{\mathrm{MD}}\right|
    }{
        \frac{1}{N}\sum_{i=1}^{N}
        f_i^{\mathrm{MD}}+\epsilon
    }.
\end{equation}
We further compare the eigenvalues of each $3\times3$ marginal block via their Pearson correlation
\begin{equation}
    r_{\mathrm{eig}}
    =
    \frac{
        \operatorname{cov}
        \left(
        \{\hat{\lambda}_{i,j}\},
        \{\lambda^{\mathrm{MD}}_{i,j}\}
        \right)
    }{
        \sigma(\{\hat{\lambda}_{i,j}\})\,
        \sigma(\{\lambda^{\mathrm{MD}}_{i,j}\})
    },
\end{equation}
where $\lambda_{i,j}$ denotes the $j$-th eigenvalue of residue $i$'s marginal $3\times3$ block,
and the correlation is computed over all $3N$ eigenvalues stacked across the protein.
This measures recovery of the directional variance structure of each residue's local fluctuations.
For the per-target RMSF metric, we compute the Pearson correlation
$r_{\mathrm{RMSF}}^{(p)}$ between the predicted and MD $\{f_i\}$ within each protein $p$,
and report the mean across all $P$ targets:
\begin{equation}
    \text{Per-target RMSF } r
    =
    \frac{1}{P}
    \sum_{p=1}^{P}
    r_{\mathrm{RMSF}}^{(p)}.
\end{equation}
\paragraph{Residue coupling.}
For each residue pair $(i,j)$, we compute the dynamic cross-correlation
\begin{equation}
    C_{ij}
    =
    \frac{
        \operatorname{tr}(\Sigma_{ij})
    }{
        \sqrt{
            \operatorname{tr}(\Sigma_{ii})
            \operatorname{tr}(\Sigma_{jj})
        }+\epsilon
    }.
\end{equation}
We report Pearson correlations between predicted and simulated coupling matrices over all nonlocal pairs satisfying $|i-j|>3$.
Long-range coupling is evaluated on the subset of nonlocal pairs whose frame-0 C$\alpha$ distance exceeds $8\,\text{\AA}$.
\paragraph{Collective dynamics.}
We eigendecompose the predicted and simulated joint matrices as
\begin{equation}
    \hat{\Sigma}
    =
    \hat{U}\hat{\Lambda}\hat{U}^{\top},
    \qquad
    \Sigma_{\mathrm{MD}}
    =
    U_{\mathrm{MD}}\Lambda_{\mathrm{MD}}U_{\mathrm{MD}}^{\top}.
\end{equation}
Agreement between the leading $k$-dimensional collective-motion subspaces is measured by
\begin{equation}
    S_k
    =
    \frac{1}{k}
    \left\|
        \hat{U}_k^{\top}U_{\mathrm{MD},k}
    \right\|_F^2,
\end{equation}
where $k=20$ for the top-20 subspace overlap reported in the main results.
$S_{20}$ is computed per protein and then averaged across all evaluated proteins.
\paragraph{Ensemble-level Wasserstein distances.}
We evaluate distributional agreement between the generated and MD conformational
ensembles using per-residue Gaussian Wasserstein-2 distances. For each residue $i$,
let $\mu_i,\widehat\Sigma_{ii}$ and $\mu_i^{\mathrm{MD}},\Sigma_{ii}^{\mathrm{MD}}$
denote the mean position and $3\times3$ marginal covariance estimated from
the generated and MD ensembles, respectively. Let
\(
d_{\mathrm{Bures}}^2(A,B)
=
\operatorname{tr}\!\bigl(A+B-2(A^{1/2}BA^{1/2})^{1/2}\bigr)
\)
denote the squared Bures metric between two $3\times3$ covariance matrices.
The Gaussian W2 distance for residue $i$ is then
\begin{equation}
    W_{2,i}^2
    =
    \bigl\|
    \mu_i-\mu_i^{\mathrm{MD}}
    \bigr\|_2^2
    +
    d_{\mathrm{Bures}}^2
    \bigl(
    \widehat\Sigma_{ii},\,
    \Sigma_{ii}^{\mathrm{MD}}
    \bigr).
\end{equation}
The root-mean W2 distance aggregates these per-residue distances:
\begin{equation}
    \text{Root-mean } W_2
    =
    \sqrt{
    \frac{1}{N}
    \sum_{i=1}^{N}
    W_{2,i}^2
    }.
\end{equation}
It decomposes additively into translation and variance components:
\begin{equation}
    (\text{Root-mean }W_2)^2
    =
    (\text{Translation }W_2)^2
    +
    (\text{Variance }W_2)^2,
\end{equation}
where
\begin{equation}
    \text{Translation }W_2
    =
    \sqrt{
    \frac{1}{N}
    \sum_{i=1}^{N}
    \bigl\|
    \mu_i-\mu_i^{\mathrm{MD}}
    \bigr\|_2^2
    },
\end{equation}
and
\begin{equation}
    \text{Variance }W_2
    =
    \sqrt{
    \frac{1}{N}
    \sum_{i=1}^{N}
    d_{\mathrm{Bures}}^2
    \bigl(
    \widehat\Sigma_{ii},\,
    \Sigma_{ii}^{\mathrm{MD}}
    \bigr)
    },
\end{equation}
Reported values are the mean across all evaluated proteins.
We further compute the uniform (empirical) Wasserstein-2 distance in the
top-2 principal-component space. MD-PCA~W2 fits PCA on the MD reference
ensemble and projects both ensembles onto the leading two PCs; Joint-PCA~W2
fits PCA on the concatenated MD--generated ensemble and projects both
ensembles onto the resulting top two PCs. In both cases, both
ensembles are subsampled to a common size, and the exact W2 distance is
obtained by optimal matching via the Hungarian algorithm.
\paragraph{Contact-based observables.}
A residue pair $(i,j)$ with $|i-j|>3$ is defined to be in contact in a given
frame if its C$\alpha$ distance falls below $d_{\mathrm{c}}=8\,\text{\AA}$.
For each pair, the contact probability $p_{ij}$ is the fraction of frames in
which this condition holds. Let $n_{ij}=1$ if the reference-structure
C$\alpha$ distance is below $d_{\mathrm{c}}$, and $0$ otherwise.
Weak contacts are native pairs ($n_{ij}=1$) whose contact probability is
below $0.9$:
\begin{equation}
    \mathcal{W}
    =
    \bigl\{
    (i,j)
    \;|\;
    n_{ij}=1
    \;\land\;
    p_{ij}<0.9
    \bigr\}.
\end{equation}
Transient contacts are non-native pairs ($n_{ij}=0$) whose contact probability
exceeds $0.1$:
\begin{equation}
    \mathcal{T}
    =
    \bigl\{
    (i,j)
    \;|\;
    n_{ij}=0
    \;\land\;
    p_{ij}>0.1
    \bigr\}.
\end{equation}
Agreement between the predicted and MD sets of weak (resp.\ transient) contacts
is measured by the Jaccard index
\begin{equation}
    J(\mathcal{A},\mathcal{B})
    =
    \frac{
        |\mathcal{A}\cap\mathcal{B}|
    }{
        |\mathcal{A}\cup\mathcal{B}|
    },
\end{equation}
and the mean across all evaluated proteins is reported.
\paragraph{ANM absolute scale calibration.}
ProDy ANM normal-mode variances are defined only up to an arbitrary multiplicative constant; their raw units are not directly interpretable as physical displacement amplitudes.
Scale-invariant quantities---correlation coefficients, normalized eigenvalue spectra, dynamic cross-correlations, and eigenspace overlap---are therefore directly comparable without further calibration.
For absolute metrics (RMS displacement NMAE, Root-mean $W_2$, Variance $W_2$, and the MD-PCA and Joint-PCA $W_2$ distances), a per-protein global scale factor $\kappa$ is applied to the ANM covariance before metric computation:
\begin{equation}
    \kappa
    =
    \left(
    \frac{1.0}
    {\bar{f}_{\mathrm{ANM}}^{\mathrm{raw}}}
    \right)^2,
\end{equation}
where
\(
\bar{f}_{\mathrm{ANM}}^{\mathrm{raw}}
=
\frac{1}{N}\sum_{i=1}^{N}
\sqrt{\operatorname{tr}(\Sigma_{ii}^{\mathrm{ANM,raw}})}
\)
is the mean per-residue RMSF computed from the unscaled ANM modes.
Residue displacements drawn from the scaled covariance
$\Sigma^{\mathrm{ANM}} = \kappa\,\Sigma^{\mathrm{ANM,raw}}$
then have a mean RMSF of $1.0\,\text{\AA}$.
This auto-scaling uses only the input backbone geometry; it does not access MD reference data and is applied independently to every evaluated protein.
All ANM absolute quantities reported in the main text incorporate this calibration.
\section{Error Decomposition and Empirical Validation of the Structured Covariance Representation}
\begin{figure*}[t]
\centering
\includegraphics[width=1.9\columnwidth]{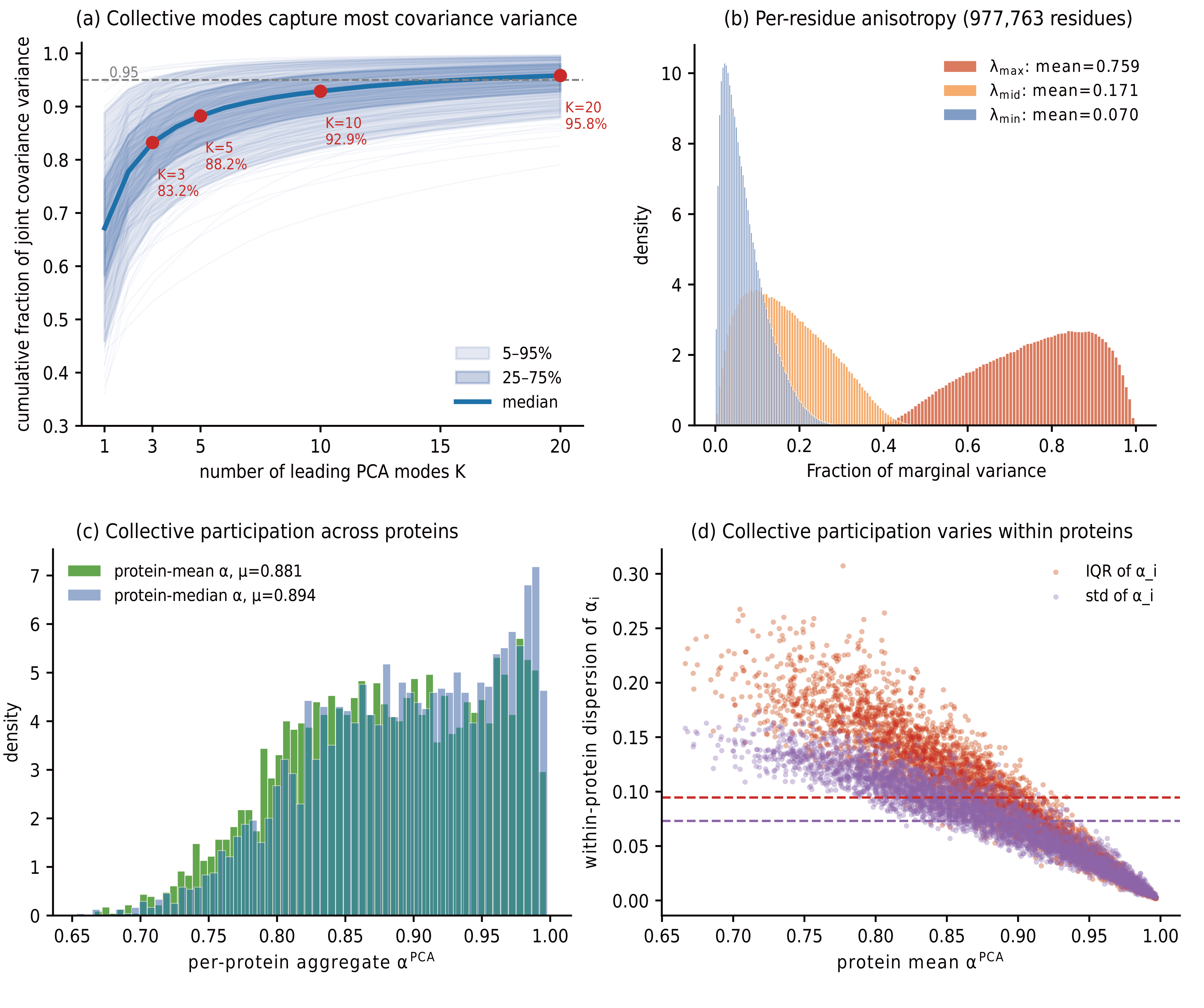}
\caption{Empirical analysis for the structured covariance decomposition on ATLAS.
(a) Cumulative fraction of empirical covariance variance explained by the leading PCA modes across protein trajectories.
(b) Distribution of the fractions of residue-wise marginal variance explained by the largest, intermediate, and smallest eigenvalues.
(c) Distribution of protein-level mean and median collective participation.
(d) Within-protein dispersion of residue-wise collective participation as a function of protein-level mean participation. Dashed lines indicate dataset-level mean IQR and standard deviation.}
\label{si:fig1}
\end{figure*}
We analyze the approximation error introduced by the proposed structured
covariance representation. Rather than directly approximating the full
joint covariance matrix, the proposed decomposition preserves residue-wise
marginal statistics exactly and introduces approximations only in the
representation of correlated motion. Specifically, the total covariance is
decomposed into a collective component and a residue-local residual:
\begin{equation}
\Sigma^\star
=
\Sigma^\star_{\mathrm{coll}}
+
\Sigma^\star_{\mathrm{res}},
\end{equation}
where $\Sigma^\star$ denotes the empirical covariance estimated from MD
trajectories. The proposed representation approximates this covariance as
\begin{equation}
\hat{\Sigma}
=
\hat{\Sigma}_{\mathrm{coll}}
+
\hat{\Sigma}_{\mathrm{res}},
\end{equation}
where $\hat{\Sigma}_{\mathrm{coll}}=BB^\top$ is a rank-$K$ collective
component and $\hat{\Sigma}_{\mathrm{res}}$ is the residue-local residual.
The representation error can therefore be attributed to two sources:
\begin{equation}
\left\|
\Sigma^\star-\hat{\Sigma}
\right\|
\leq
\underbrace{
\left\|
\Sigma^\star_{\mathrm{coll}}
-
\hat{\Sigma}_{\mathrm{coll}}
\right\|
}_{\text{collective approximation}}
+
\underbrace{
\left\|
\Sigma^\star_{\mathrm{res}}
-
\hat{\Sigma}_{\mathrm{res}}
\right\|
}_{\text{variance approximation}}.
\end{equation}
The following sections analyze these two approximation sources
independently.
\begin{table*}[h]
\centering
\small
\setlength{\tabcolsep}{4pt}
\renewcommand{\arraystretch}{1.05}
\begin{tabular}{@{}llccc@{}}
\toprule
Category & Reconstruction & Region & Mean $\pm$ Std & p05 / p95 \\
\midrule
\multirow{2}{*}{Unconstrained low-rank reference}
& PCA20 raw
& full
& 0.009 $\pm$ 0.007
& 0.001 / 0.023
\\
& PCA20 + exact block tail
& full
& 0.008 $\pm$ 0.006
& 0.001 / 0.019
\\
\midrule
\multirow{1}{*}{Structured oracle}
& Structured + true $\alpha_i$
& full
& 0.037 $\pm$ 0.023
& 0.006 / 0.078
\\
\midrule
\multirow{2}{*}{Participation ablation}
& Structured + $\alpha_i=1$
& full
& 0.046 $\pm$ 0.033
& 0.006 / 0.109
\\
& Structured + mean $\bar{\alpha}$
& full
& 0.099 $\pm$ 0.059
& 0.013 / 0.201
\\
\bottomrule
\end{tabular}
\caption{
Oracle analysis of representation on ATLAS.
Relative Frobenius errors are computed against empirical joint
covariances.
PCA-based reconstructions provide unconstrained low-rank references,
whereas structured variants quantify the approximation introduced by
the proposed physics-based parameterization and its ablations.
}
\label{tab:oracle_floor}
\end{table*}
\paragraph{Collective approximation.}
The collective component is represented by a rank-$K$ positive
semidefinite matrix. Let the empirical collective covariance admit the
eigendecomposition
\begin{equation}
\Sigma^\star_{\mathrm{coll}}
=
U\Lambda U^\top,
\end{equation}
where the eigenvalues are ordered as
\[
\lambda_1\geq\lambda_2\geq\cdots\geq\lambda_{3N}\geq0.
\]
The optimal rank-$K$ approximation under the Frobenius norm is given by
\begin{equation}
\Sigma^{(K)}_{\mathrm{coll}}
=
U_K\Lambda_KU_K^\top .
\end{equation}
According to the Eckart--Young--Mirsky theorem, the minimum achievable
error among all rank-$K$ approximations is
\begin{equation}
\min_{\mathrm{rank}(A)\leq K}
\left\|
\Sigma^\star_{\mathrm{coll}}-A
\right\|_F
=
\left(
\sum_{i>K}\lambda_i^2
\right)^{1/2}.
\end{equation}
Therefore, the irreducible error introduced by collective subspace
truncation is determined solely by the discarded covariance spectrum.
The corresponding normalized error is
\begin{equation}
\epsilon_K
=
\frac{
\left(\sum_{i>K}\lambda_i^2\right)^{1/2}
}{
\left(\sum_i\lambda_i^2\right)^{1/2}
}.
\end{equation}
This establishes that the approximation error of the collective component
is controlled by the spectral decay of the covariance matrix rather than
the parameterization itself.
\paragraph{Exact marginal preservation.}
In contrast to the collective component, residue-wise marginal covariance
is preserved exactly. By construction,
\begin{equation}
\left[
\hat{\Sigma}_{\mathrm{coll}}
\right]_{ii}
=
\alpha_i\Sigma_{\mathrm{marg},i},
\end{equation}
and
\begin{equation}
\hat{\Sigma}_{\mathrm{res},i}
=
(1-\alpha_i)\Sigma_{\mathrm{marg},i}.
\end{equation}
Therefore,
\begin{equation}
\left[
\hat{\Sigma}
\right]_{ii}
=
\hat{\Sigma}_{\mathrm{coll},ii}
+
\hat{\Sigma}_{\mathrm{res},i}
=
\Sigma_{\mathrm{marg},i}.
\end{equation}
Hence, the proposed representation introduces no approximation error for
residue-wise fluctuation statistics, including local fluctuation magnitude
and anisotropy. The remaining approximation originates only from modeling
the correlated component.
\paragraph{Variance approximation.}
The remaining approximation arises from the assumption that the residue-wise
contribution to collective motion can be represented through a scalar
participation coefficient. Specifically, the collective variance allocated
to residue $i$ is modeled as
\begin{equation}
\hat{\Sigma}_{\mathrm{coll},ii}
=
\alpha_i\Sigma_{\mathrm{marg},i}.
\end{equation}
For an empirical covariance, the exact collective contribution may not be
strictly proportional to the marginal covariance. The deviation from this
scalar allocation assumption can therefore be characterized as
\begin{equation}
\epsilon_{\alpha}
=
\sum_i
\left\|
\Sigma^\star_{\mathrm{coll},ii}
-
\alpha_i\Sigma^\star_{\mathrm{marg},i}
\right\|_F .
\end{equation}
Unlike collective truncation, this term does not arise from discarded
dimensions but from enforcing an interpretable and controllable
factorization of residue-wise collective participation.
\paragraph{Empirical validation.}
We evaluate the above approximation sources on ATLAS dataset. The covariance spectrum is strongly
concentrated in a low-dimensional collective subspace. The median cumulative
variance explained by the leading five and twenty modes is 0.882 and 0.958,
respectively, indicating that the collective truncation error is small.
Residue-wise marginal covariance is strongly anisotropic. Across all
residue-level covariance blocks, the largest eigenvalue explains 75.9\% of
local variance on average, whereas the smallest eigenvalue explains only
7.0\%, demonstrating the necessity of retaining full three-dimensional
marginal covariance rather than scalar flexibility measures.
Finally, we quantify the overall approximation floor using oracle quantities
extracted from empirical covariance (Table~\ref{tab:oracle_floor}). Using
empirical marginal covariance, top-20 collective factors, and
residue-specific participation achieves a mean relative Frobenius error of
0.037. This error is only moderately larger than the unconstrained PCA20
reference (0.009), reflecting the cost of enforcing exact marginal
consistency and factor disentanglement. In contrast, replacing
residue-specific participation with a protein-wise mean increases the error
to 0.099, confirming the importance of residue-level participation.
\section{Numerical Stability of the Covariance Assembly}
\label{sec:supp_numerical_stability}
\begin{table}[t]
\centering
\small
\setlength{\tabcolsep}{3pt}
\begin{tabular}{lccc}
\toprule
Block & Operation & $\epsilon$ & Triggered\\
\midrule
$G_i$
& inv. sqrt.
& $10^{-6}$
& 0/10000
\\
$\alpha_i\Sigma_i$
& sqrt.
& $10^{-8}$
& 0/10000
\\
$\Sigma_i$
& PSD check
& --
& 0/10000
\\
$\lambda$
& clipping
& $10^{-8}$
& 0/1000
\\
\bottomrule
\end{tabular}
\caption{
Numerical safeguard audit during inference.
}
\label{tab:eig_floor}
\end{table}
The proposed fractional-gating covariance assembler constructs
$\Sigma_{\mathrm{joint}}\in\mathbb{S}^{3N}_{++}$ through a sequence of
matrix square-root and inverse-square-root operations. Although the
parameterization is designed to preserve positive definiteness under exact
arithmetic, finite-precision computation may introduce numerical issues when
intermediate covariance blocks become ill-conditioned. Therefore, we introduce
eigenvalue floors as numerical safeguards during inference.
Specifically, three operations are protected:
the inverse square root of the row Gram matrix
$G_i=V^{\mathrm{raw}}_i(V^{\mathrm{raw}}_i)^{\top}$,
the square root of the allocated collective variance
$T_i=\alpha_i\Sigma_{\mathrm{marg},i}$,
and the mode-level eigenvalues $\lambda$ used in spectral operations.
The corresponding numerical safeguards are summarized in
Table~\ref{tab:eig_floor}.
\paragraph{Numerical stability evaluation.}
To quantify whether these numerical safeguards are activated in practice, we
replayed the complete covariance assembly procedure using float64 arithmetic
on the production inference sweep. The analysis contains 50 generated proteins
spanning lengths
$\{100,150,200,250,300\}$, with 10 proteins per length.
The evaluated quantities include all intermediate covariance blocks and
spectral components involved in the assembly process, including predicted
collective directions
$V_{\mathrm{pred}}\in\mathbb{R}^{3N\times20}$,
mode eigenvalues, residue-wise participation coefficients
$\alpha_i$, and marginal covariance blocks
$\Sigma_{\mathrm{marg},i}$.
For each residue, we recorded whether the minimum eigenvalue of each
intermediate matrix fell below the corresponding numerical threshold. In total,
the analysis covers 10,000 residue blocks and 1,000 collective modes.
\paragraph{Conditioning of intermediate matrices.}
Although no numerical floor was activated, we further examined the distance
between the intermediate spectra and the applied thresholds. The first
percentile of minimum eigenvalues across different protein lengths is reported
in Table~\ref{tab:conditioning}.
\begin{table}[t]
\centering
\small
\setlength{\tabcolsep}{3pt}
\begin{tabular}{ccccc}
\toprule
Length
& $G_i$
& $\alpha_i\Sigma_i$
& $\Sigma_i$
& Mode
\\
\midrule
100
& $9.44e{-3}$
& $1.33e{-1}$
& $1.57e{-1}$
& $4.57e{-1}$
\\
150
& $5.53e{-3}$
& $1.22e{-1}$
& $1.60e{-1}$
& $4.03e{-1}$
\\
200
& $4.04e{-3}$
& $1.42e{-1}$
& $1.54e{-1}$
& $3.81e{-1}$
\\
250
& $3.03e{-3}$
& $1.45e{-1}$
& $1.67e{-1}$
& $3.69e{-1}$
\\
300
& $2.22e{-3}$
& $1.37e{-1}$
& $1.62e{-1}$
& $3.60e{-1}$
\\
\bottomrule
\end{tabular}
\caption{
First percentile of minimum eigenvalues across protein lengths.
Columns correspond to $G_i$, $\alpha_i\Sigma_i$,
$\Sigma_i$, and mode eigenvalues.
}
\label{tab:conditioning}
\end{table}
The row-Gram spectrum exhibits a mild length dependence, with the first
percentile of $\lambda_{\min}(G_i)$ decreasing from
$9.4\times10^{-3}$ at length 100 to
$2.2\times10^{-3}$ at length 300. Nevertheless, these values remain more than
three orders of magnitude above the applied floor of $10^{-6}$.
The remaining numerical safeguards exhibit substantially larger margins.
The allocated collective covariance blocks remain well-conditioned, with
minimum eigenvalues around $1.5\times10^{-1}$, while the mode-level
eigenvalues remain within approximately
$[3.6\times10^{-1},4.6\times10^{-1}]$, far above the
$10^{-8}$ clipping threshold.
These results demonstrate that eigenvalue floors function only as theoretical
numerical backstops and are not activated during the evaluated inference
runs. Therefore, the reported covariance statistics reflect the learned
fractional-gating representation rather than contributions from numerical
clipping or stabilization procedures.
\section{Additional Evaluation Metrics}
\begin{table*}[htbp]
  \centering
  \small
  \begin{tabular}{c c c c c c}
  \toprule
  \textbf{Length}
  & \textbf{BioEmu}
  & \textbf{MDGen}
  & \textbf{Str2Str}
  & \multicolumn{2}{c}{\textbf{PhiFold}}
  \\
  \cmidrule(lr){5-6}
  &
  (ms / conf.)
  &
  (ms / conf.)
  &
  (ms / conf.)
  &
  end-to-end amortised
  &
  ensemble expansion
  \\
  &
  &
  &
  &
  (ms / conf., $K{=}1000$)
  &
  ($\mu$s / conf.)
  \\
  \midrule
  100
  & 529.286 $\pm$ 215.205
  & 61.272 $\pm$ 8.027
  & 13998.347 $\pm$ 4224.408
  & \textbf{1.91 $\pm$ 0.05}
  & \textbf{0.311 $\pm$ 0.010}
  \\
  150
  & 846.673 $\pm$ 389.510
  & 75.463 $\pm$ 22.723
  & 14782.796 $\pm$ 4467.001
  & \textbf{1.89 $\pm$ 0.04}
  & \textbf{0.404 $\pm$ 0.010}
  \\
  200
  & 790.854 $\pm$ 33.287
  & 112.116 $\pm$ 19.152
  & 16522.789 $\pm$ 4985.198
  & \textbf{1.90 $\pm$ 0.02}
  & \textbf{0.443 $\pm$ 0.004}
  \\
  250
  & 1211.111 $\pm$ 36.525
  & 76.746 $\pm$ 9.747
  & 18784.649 $\pm$ 5625.874
  & \textbf{1.90 $\pm$ 0.01}
  & \textbf{0.496 $\pm$ 0.012}
  \\
  300
  & 2616.633 $\pm$ 701.803
  & 92.713 $\pm$ 13.573
  & 21312.320 $\pm$ 6463.117
  & \textbf{1.97 $\pm$ 0.02}
  & \textbf{0.619 $\pm$ 0.054}
  \\
  \bottomrule
  \end{tabular}
  \vspace{4pt}
  \begin{minipage}{\textwidth}
\footnotesize
PhiFold end-to-end cost is amortised over $K{=}1000$ conformers:
\[
\bigl(t_{\mathrm{backbone}+\Sigma_{\mathrm{joint}}}
+t_{\mathrm{expansion}}(K)\bigr)/K.
\]
Ensemble expansion is measured from precomputed
$\Sigma_{\mathrm{joint}}$ tensors.
\end{minipage}
\caption{
Inference cost for generating 1000-frame ensembles.
PhiFold reports the end-to-end amortised cost and the incremental
ensemble-expansion cost after covariance prediction.
}
\label{tab:efficiency}
\end{table*}
\subsection{Inference Efficiency}
We benchmark the inference efficiency of equilibrium ensemble generation
across five protein length categories (100--300 residues).
All methods generate 1000 conformers per protein on a single NVIDIA H100
80\,GB HBM3 GPU. Reported values correspond to the mean wall-clock
runtime amortised over the generated conformers.
For BioEmu, all sequence-dependent preprocessing required for ensemble
generation is included in the reported runtime and amortised over the
generated conformers. For MDGen, we generate a 1000-frame trajectory
rollout conditioned on the input structure. For Str2Str, we follow its
inference protocol and report the reverse-diffusion sampling cost with
structure representations prepared prior to sampling.
Table~\ref{tab:efficiency} summarizes the inference cost of different
ensemble-generation strategies. Sampling-based approaches generate
ensemble members through repeated neural generation steps, resulting in
approximately linear growth of computational cost with the number of
generated conformers. In contrast, PhiFold predicts the backbone and
equilibrium covariance representation once, after which additional
conformers are generated through covariance-based stochastic expansion.
We report two efficiency measurements for PhiFold. The end-to-end
amortised cost includes the one-time backbone-$+\Sigma_{\mathrm{joint}}$
prediction cost distributed over 1000 generated conformers. The ensemble
expansion cost measures the incremental cost of generating additional
conformers after $\Sigma_{\mathrm{joint}}$ has been constructed.
PhiFold substantially reduces the amortised cost of ensemble generation
across all tested protein lengths. At $K=1000$, the fixed covariance
prediction overhead is amortised over the generated ensemble, while
additional conformers can be sampled at negligible incremental cost.
This efficiency advantage arises from replacing repeated neural sampling
with a single prediction of an equilibrium covariance representation
followed by efficient covariance-based ensemble expansion.
\subsection{Designability and Diversity}
The backbone generation network is inherited from FoldFlow and remains frozen throughout training. We therefore expect the designability and structural diversity of the main-result samples to remain close to those of FoldFlow. The empirical results are consistent with this expectation: among the 250 evaluated structures, 80.8\% satisfy the designability criterion.
The mean pairwise TM-score is 0.377, indicating substantial structural diversity and remaining close to the FoldFlow result. These observations suggest that introducing the dynamics-prediction component does not substantially alter the generative behavior of the frozen backbone model.
\begin{table}[H]
\centering
\small
\setlength{\tabcolsep}{6pt}
\begin{tabular}{lc}
\toprule
Metric & Value \\
\midrule
Designability & 80.8\% \\
Mean maximum scTM & 0.919 \\
Mean minimum RMSD (\AA) & 1.785 \\
Diversity (mean pairwise TM-score) & 0.377 \\
\bottomrule
\end{tabular}
\caption{
Aggregate evaluation over 250 generated structures pooled across all
protein lengths.
}
\label{tab:main_additional_metrics}
\end{table}
\subsection{Structural validity of generated ensembles}
To assess whether covariance-based ensemble expansion introduces structural
artifacts, we evaluated the geometric validity of both generated backbone
structures and sampled conformational ensembles. We followed the
C$\alpha$-C$\alpha$ distance validity criterion used in FoldFlow, where a
consecutive C$\alpha$ pair is considered valid if its distance is within
0.1~\AA{} above the ideal distance. We further evaluated
atom clashes, backbone stereochemistry, and global compactness.
As shown in Table~\ref{tab:structural_validity}, generated backbones exhibit
high structural quality, satisfying the bond distance criterion and
maintaining favorable backbone geometry. After ensemble expansion, sampled
conformations preserve similar structural characteristics while exhibiting
the expected increase in conformational variability. The comparable global
compactness between generated backbones and sampled ensembles indicates that
covariance-based sampling explores alternative conformations without causing
unphysical unfolding or structural collapse.
\begin{table}[t]
\centering
\small
\begin{tabular}{lcc}
\toprule
Metric & Backbone & Ensemble \\
\midrule
C$\alpha$ distance validity (\%)
& $100.00 \pm 0.00$
& $92.03 \pm 1.91$ \\
C$\alpha$--C$\alpha$ distance std. (\AA)
& $0.155 \pm 0.003$
& $0.486 \pm 0.077$ \\
Steric validity (\%)
& $99.85 \pm 0.18$
& $98.66 \pm 0.96$ \\
Angle favored (\%)
& $98.67 \pm 0.22$
& 84.92 $\pm$ 3.30\\
$R_g/L^{1/3}$
& $3.030 \pm 0.061$
& $3.035 \pm 0.211$ \\
\bottomrule
\end{tabular}
\caption{
Structural validity before and after covariance-based ensemble expansion.
C$\alpha$ distance validity is computed following FoldFlow with a 0.1~\AA{}
tolerance above the ideal distance.
}
\label{tab:structural_validity}
\end{table}
\begin{figure}[H]
\centering
\includegraphics[width=0.9\columnwidth]{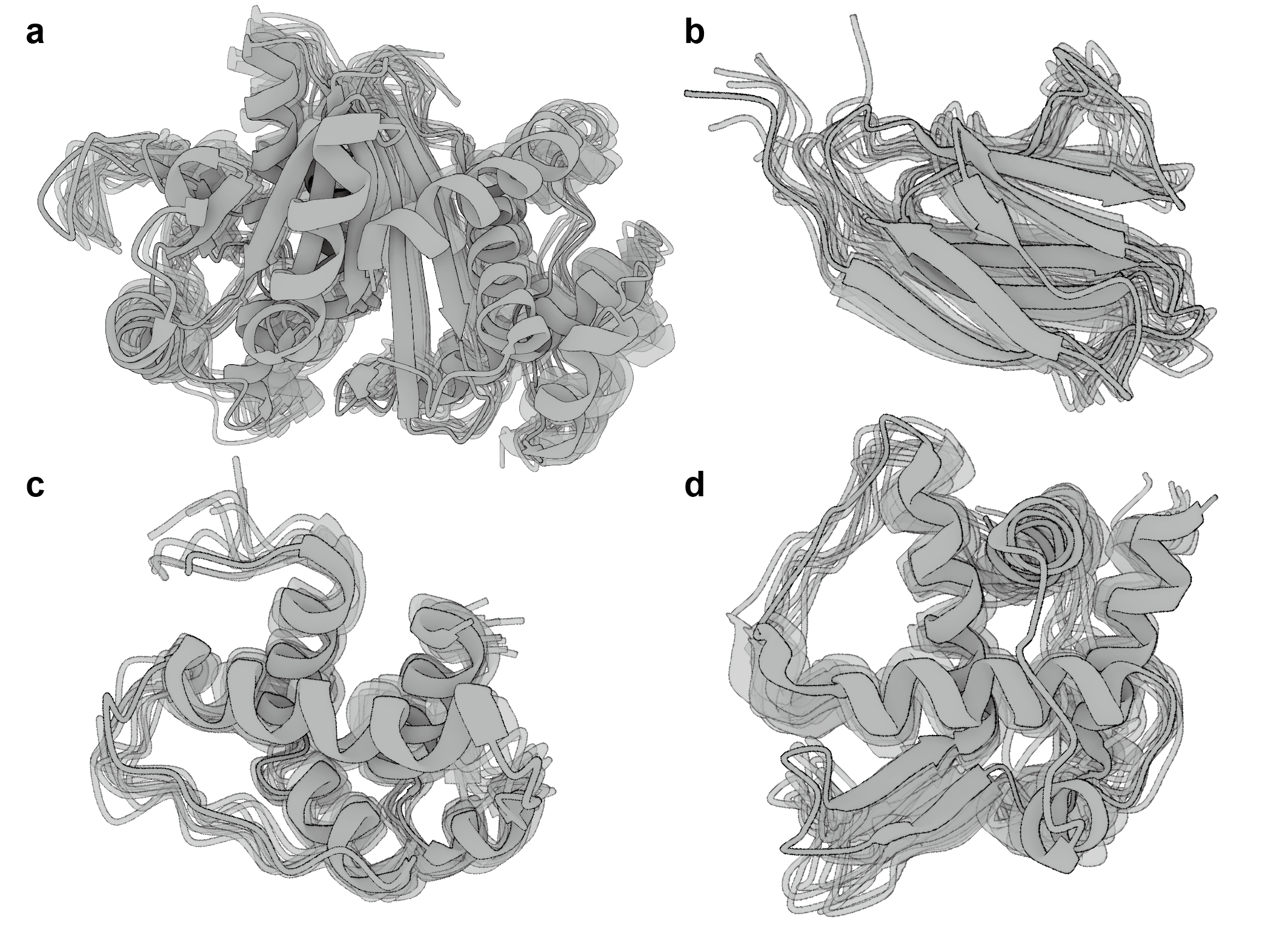}
\caption{Visualization of generated structural ensemble. 10 conformational samples from each predicted ensemble are aligned and superimposed.}
\label{si:fig2}
\end{figure}
\section{Backbone-Level Effects and Designability under Regional Guidance}
\label{sec:guidance_backbone_analysis}
\begin{table*}[h]
\centering
\small
\begin{tabular}{llccccccc}
\toprule
Length
& Condition
& \shortstack{Target\\bbRMSD}
& \shortstack{Outside\\bbRMSD}
& \shortstack{Localization\\ratio}
& $\Delta$Ordered
& \shortstack{ANM RMSIP\\top-20}
& \shortstack{Design-\\ability}
& \shortstack{Mean max\\scTM} \\
&
&
& (\AA)
& (\AA)
&
& (pp)
&
& (\%)
\\
\midrule
\multirow{3}{*}{$100^{\dagger}$}
& Uncond.
& --
& --
& --
& --
& --
& 100
& 0.957 \\
& Flex
& $5.850 \pm 6.509$
& $2.089 \pm 2.619$
& 2.80
& $+0.13$
& $0.827 \pm 0.144$
& 100
& 0.956 \\
& Rigid
& $1.954 \pm 1.654$
& $0.766 \pm 1.123$
& 2.55
& $+0.63$
& $0.914 \pm 0.079$
& 100
& 0.948 \\
\midrule
\multirow{3}{*}{150}
& Uncond.
& --
& --
& --
& --
& --
& 100
& 0.957 \\
& Flex
& $4.271 \pm 2.864$
& $2.340 \pm 1.247$
& 1.83
& $-0.75$
& $0.837 \pm 0.097$
& 95
& 0.945 \\
& Rigid
& $4.450 \pm 4.702$
& $1.805 \pm 1.267$
& 2.46
& $+1.25$
& $0.855 \pm 0.120$
& 100
& 0.954 \\
\midrule
\multirow{3}{*}{200}
& Uncond.
& --
& --
& --
& --
& --
& 80
& 0.929 \\
& Flex
& $5.928 \pm 2.329$
& $3.132 \pm 1.758$
& 1.89
& $-2.00$
& $0.812 \pm 0.083$
& 80
& 0.929 \\
& Rigid
& $4.043 \pm 4.104$
& $2.327 \pm 2.038$
& 1.74
& $+5.38$
& $0.865 \pm 0.107$
& 85
& 0.927 \\
\midrule
\multirow{3}{*}{250}
& Uncond.
& --
& --
& --
& --
& --
& 75
& 0.905 \\
& Flex
& $6.027 \pm 2.636$
& $3.509 \pm 1.946$
& 1.72
& $-2.75$
& $0.807 \pm 0.083$
& 65
& 0.873 \\
& Rigid
& $3.850 \pm 3.086$
& $2.550 \pm 1.855$
& 1.51
& $+1.75$
& $0.863 \pm 0.084$
& 75
& 0.899 \\
\midrule
\multirow{3}{*}{300}
& Uncond.
& --
& --
& --
& --
& --
& 20
& 0.735 \\
& Flex
& $7.261 \pm 3.028$
& $3.952 \pm 1.578$
& 1.84
& $-2.63$
& $0.819 \pm 0.070$
& 25
& 0.756 \\
& Rigid
& $4.294 \pm 2.473$
& $2.963 \pm 1.617$
& 1.45
& $+3.38$
& $0.873 \pm 0.067$
& 25
& 0.671 \\
\bottomrule
\end{tabular}
\caption{
Backbone-level effects and designability under regional guidance.
Backbone displacement, localization, $\Delta$Ordered, and ANM RMSIP
are defined relative to matched unconditional samples and are therefore
not applicable to the unconditional rows.
Strict designability is the percentage passing the strict designability
criterion, and scTM is the mean maximum self-consistency TM-score.
$^{\dagger}$Length-100 regional RMSDs use
$\texttt{min\_outside\_frac}=0.15$ and a 20-residue outside-region
Kabsch anchor, so their regional decomposition should be interpreted
cautiously.
}
\label{tab:guidance_backbone_designability}
\end{table*}
We next examine whether regional guidance affects the generated backbone and its designability. For each guided sample and its matched unconditional counterpart, we compute the C$\alpha$ RMSD within and outside the selected region, the change in ordered secondary structure, and the top-20 ANM subspace overlap. Designability is assessed using the strict success rate and mean maximum scTM, with unconditional samples included as the reference.
As shown in Table~\ref{tab:guidance_backbone_designability}, the target-region RMSD exceeds the outside-region RMSD across all evaluated guided conditions, indicating preferential structural displacement within the selected region. The length-100 regional decomposition is interpreted cautiously because of its small outside-region alignment anchor and is not used to infer a length-dependent localization trend.
Flexibility guidance generally reduces ordered secondary structure, whereas rigidity guidance increases it. At the same time, the high ANM subspace overlaps indicate that the broader collective-motion organization remains largely preserved despite these regional changes.
Guided and unconditional samples show broadly comparable designability, with protein length producing a larger effect than guidance condition. Designability decreases for longer proteins under all conditions, while regional guidance introduces only modest additional changes relative to the unconditional baseline. Together, these results show that regional guidance preferentially modifies the selected backbone region without substantially disrupting global dynamics or overall designability.

\end{document}